\documentclass{article}

\usepackage[dblblindworkshop,final]{neurips_2025}
\workshoptitle{The Second Workshop on GenAI for Health: Potential, Trust, and Policy Compliance}

\usepackage[utf8]{inputenc} 
\usepackage[T1]{fontenc}    
\usepackage{amsfonts}       
\usepackage{hyperref}       
\usepackage{url}            
\usepackage{booktabs}       
\usepackage{nicefrac}       
\usepackage{microtype}      
\usepackage{xcolor}         
\usepackage{adjustbox}
\usepackage{booktabs,makecell,multirow,threeparttable}

\usepackage{booktabs,tabularx}
\usepackage[most]{tcolorbox}
\usepackage{listings} 
\lstdefinestyle{prompt}{
  basicstyle=\ttfamily\footnotesize,
  breaklines=true, breakatwhitespace=true,
  numbers=left, numberstyle=\tiny, numbersep=6pt,
  frame=single, columns=fullflexible
}
\tcbset{colback=gray!5,colframe=gray!40,enhanced,sharp corners,
  boxrule=0.3pt,left=6pt,right=6pt,top=6pt,bottom=6pt}

\title{Count-Based Approaches Remain Strong: A Benchmark Against Transformer and LLM Pipelines on Structured EHR}

\author{%
  Jifan Gao \\
  Department of Data Science\\
  Dana-Farber Cancer Institute\\
  Boston, MA \\
  \And
  Michael Rosenthal \\
  Department of Radiology\\
  Dana-Farber Cancer Institute\\
  Boston, MA \\  
  \And
  Brian Wolpin \\
  Department of Medical Oncology\\
  Dana-Farber Cancer Institute\\
  Boston, MA \\
  \And
  Simona Cristea \thanks{Correspondence to: \texttt{scristea@ds.dfci.harvard.edu}}\\
  Department of Data Science\\
  Dana-Farber Cancer Institute\\
  Boston, MA \\
}

\begin{document}

\maketitle

\begin{abstract}
Structured electronic health records (EHR) are essential for clinical prediction. While count-based learners continue to perform strongly on such data, no benchmarking has directly compared them against more recent mixture-of-agents LLM pipelines, which have been reported to outperform single LLMs in various NLP tasks. In this study, we evaluated three categories of methodologies for EHR prediction using the EHRSHOT dataset: count-based models built from ontology roll-ups with two time bins, based on LightGBM and the tabular foundation model TabPFN; a pretrained sequential transformer (CLMBR); and a mixture-of-agents pipeline that converts tabular histories to natural-language summaries followed by a text classifier. We assessed eight outcomes using the EHRSHOT dataset. Across the eight evaluation tasks, head-to-head wins were largely split between the count-based and the mixture-of-agents methods. Given their simplicity and interpretability, count-based models remain a strong candidate for structured EHR benchmarking. The source code is available at: 
\href{https://github.com/cristea-lab/Structured_EHR_Benchmark}{https://github.com/cristea-lab/Structured\_EHR\_Benchmark}.
\end{abstract}

\section{Introduction}

Structured electronic health record (EHR) data consists of clinical information from various domains such as medical conditions and drug exposures. These records are stored with standardized medical concepts and organized into relational tables, using standardized frameworks like the Observational Medical Outcomes Partnership (OMOP) Common Data Model \citep{hripcsak2015observational}. Compared to other data modalities, such as clinical free-text notes and medical imaging, structured EHR data offer unique advantages that make them valuable in clinical outcome prediction. Standardized vocabularies and schemas enable cross-site analyses, reproducibility, and fair benchmarking \citep{arora2023value, wang2025scoping}; the structured data can scale to million-patient cohorts for efficient cohort construction and population-level training \citep{gamal2021standardized}; and the organized nature of structured data significantly simplifies adherence to regulatory and ethical guidelines \citep{tayefi2021challenges}, such as those mandated by the Health Insurance Portability and Accountability Act (HIPAA) \citep{edemekong2018health}.

Various modeling strategies have been developed for structured EHR data. Early approaches often relied on count-based representations \citep{reps2018design, khalid2021standardized}, where the occurrence or frequency of medical concepts (\textit{e.g.}, diagnoses, procedures, medications) is aggregated into feature vectors. These representations are straightforward and compatible with common machine-learning models such as logistic regression, random forests, or gradient boosting machines, and have demonstrated strong performance on various clinical prediction tasks \citep{gao2024fair, bergquist2024crowd, bergquist2023evaluation}. Despite their simplicity, count-based models largely ignore the temporal order and contextual relationships between events.

Motivated by advances in natural language processing, more recent methods treat EHR data as sequential tokens, applying transformer architectures designed to capture contextual dependencies. Transformer-based models such as Delphi \citep{shmatko2024learning}, CLMBR \citep{steinberg2021language, guo2024multi} and MOTOR \citep{steinberg2023motor} convert time-stamped medical codes into sequences and employ attention mechanisms to model longitudinal patterns in patient histories. These pretrained frameworks have shown promising performance across tasks. In addition, researchers are exploring the use of large language models (LLMs) for structured EHR prediction by converting tabular records into textual narratives \citep{wornow2024context, kim2025medrep, kirchler2025large}. This paradigm allows multimodal integration beyond free text and leverages the text understanding capabilities of LLMs. In the general text domain, \cite{wang2024mixture} proposed a mixture-of-agents (MoA) approach and demonstrated that sequentially passing the output of one LLM to another leads to improvements in performance. Based on this collaborative potential of LLMs, \cite{gao2025moma} proposed the Mixture-of-Multimodal-Agents (MoMA) for multimodal EHR data, including structured EHR. MoMA first generates concise and clinically meaningful summaries of a patient’s tabular record and then passes these summaries to a specialized text classifier for downstream prediction.

Despite rapid advances in employing LLMs for EHR data, recent evidence suggests that strong count-based tabular models remain competitive for structured EHR prediction \citep{brown2025large}, particularly when data exhibit skewed or heavy-tailed distributions, where gradient boosting machines (GBMs) often have an advantage over neural networks \citep{mcelfresh2023neural}. At the same time, LLM evaluations for clinical prediction in structured EHR data indicate that prompt-engineered single-LLM baselines do not reliably surpass classical count-based models on structured EHR tasks \citep{chen2024clinicalbench}. In addition to labeled data, the EHRSHOT initiative \citep{wornow2023ehrshot} also introduced the pretrained sequence model CLMBR \citep{steinberg2021language}, which ranks at or near the top on many EHRSHOT task groups, as well as a public leaderboard with the recorded performance of various methods on EHRSHOT tasks. However, the EHRSHOT leaderboard does not include MoA pipelines that convert tabular records to intermediate clinical summaries before classification. 

Motivated by this knowledge gap in understanding how different classes of models perform on predicting various medical outcomes from EHR data, we hereby established a benchmark that directly compares state-of-the-art count-based learners (including GBM and the prior-data-fitted TabPFN \citep{hollmann2022tabpfn}, a strong small-data tabular foundation model) with a MoA LLM architecture for structured EHR prediction on EHRSHOT. Our paper contributes through the following: 

\begin{itemize}
    \item Head-to-head comparisons across three methodology categories: (i) count-based features with GBM and with the TabPFN foundation model, (ii) CLMBR (transformer-based), and (iii) our MoA approach;
    \item Introduction of a MoA baseline and its contribution to the EHRSHOT leaderboard;
    \item Quantitative analyses to help understand the behavior of the introduced MoA approach, demonstrating that it produces intermediate summaries adapted to the task at hand.
\end{itemize}

\begin{figure}[ht!]
    \centering
    \includegraphics[width=0.60\textwidth]{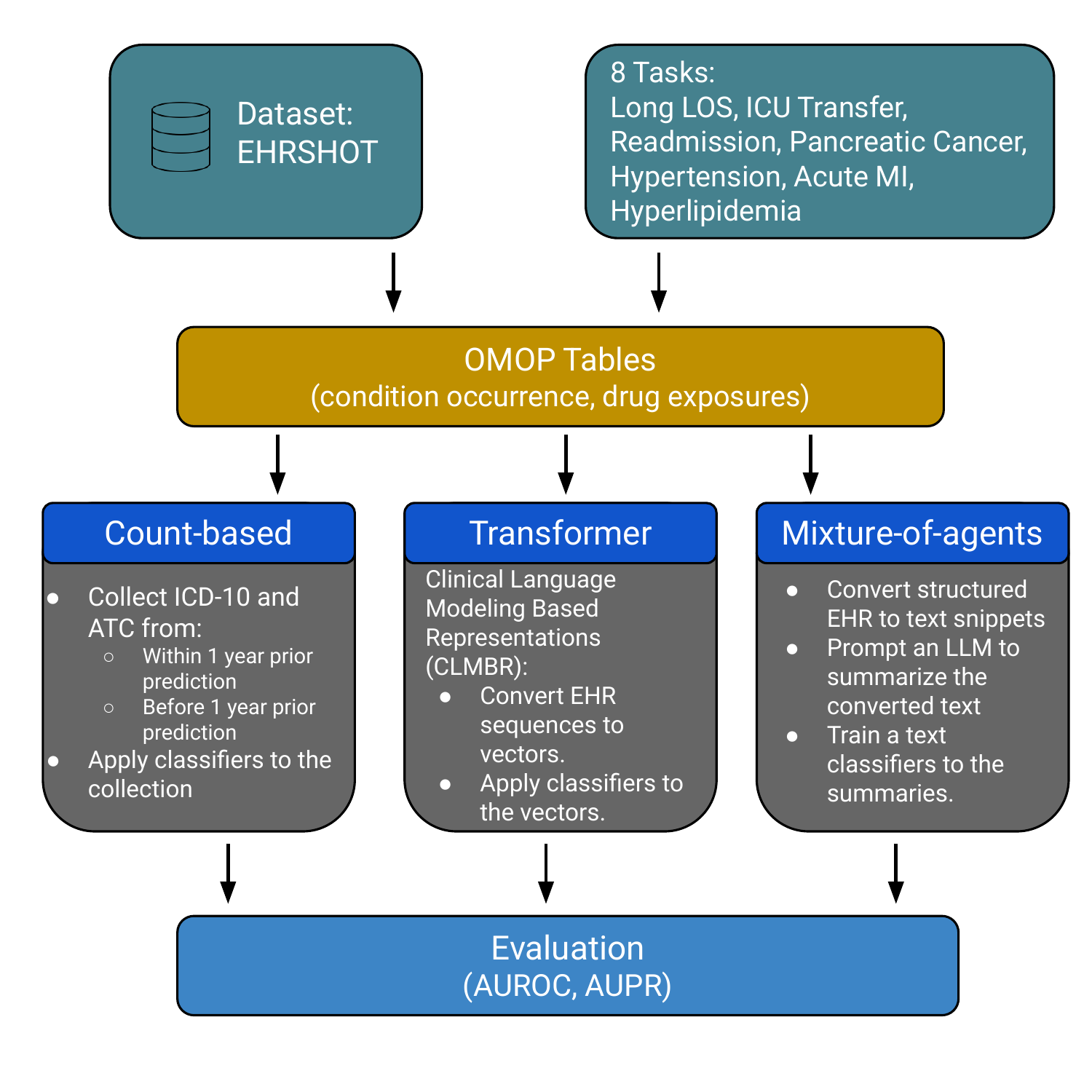}
    \caption{\textbf{Study design and benchmarking pipeline}. We evaluated three modeling categories on the EHRSHOT benchmark dataset, with OMOP-standardized cohorts and eight tasks: Long Length of Stay (LOS), ICU Transfer, Readmission, Pancreatic Cancer, Hypertension, Acute Myocardial Infarction (MI), and Hyperlipidemia. From the OMOP tables, we construct: (i) a count-based tabular pipeline that aggregates ICD-10 and ATC codes over a 1-year look-back window and trains strong tabular models (LightGBM and TabPFN); (ii) a pretrained sequential model (CLMBR) that tokenizes time-stamped OMOP concepts and learns vector representations for downstream prediction, introduced with the EHRSHOT dataset; and (iii) a MoA LLM pipeline that converts each patient’s longitudinal record into a concise natural-language summary before classification. All models use the same cohorts, prediction windows, and splits; performance is compared on held-out data.}
    \label{fig:schema}
\end{figure}

Figure \ref{fig:schema} presents the overall study design and the model schema.

\section{Methods}

In this section, we first introduce the construction of cohorts. We then describe in detail the three modeling categories used in our benchmark: (i) count-based models, which transform OMOP concepts into aggregated feature arrays for traditional machine-learning classifiers; (ii) a pretrained transformer model (CLMBR) that learns sequential representations of longitudinal patient records, introduced together with the EHRSHOT dataset; and (iii) a MoA LLM pipeline that converts tabular histories into natural-language summaries before classification.

\subsection{Cohort construction}

All analyses were conducted on the publicly available EHRSHOT benchmark, using the predefined training, validation, and test splits provided with the dataset (see \ref{tab:dataset} for cohort characteristics). Labels and prediction windows followed the definitions established in \cite{wornow2023ehrshot}. We benchmarked the three described methodology categories for eight different tasks, including three operational outcomes and five new diagnostic prediction outcomes. The operational outcomes include: Long Length of Stay (predict whether a patient’s total length of stay during a hospital visit will be at least 7 days), Readmission (predict whether a patient will be re-admitted to the hospital within 30 days after being discharged from a visit), and ICU Transfer (predict whether a patient will be transferred to the ICU during a hospital visit). The new diagnosis tasks are to predict the first diagnosis of one of these diseases within the following year after a hospital visit: pancreatic cancer, hypertension, acute myocardial infarction, hyperlipidemia, and lupus. As EHRSHOT assigns labels per hospital visit, a single patient with multiple hospital visits may contribute multiple labels. To limit bias from repeated measurements, we report the performance for predicting both the earliest assigned label per patient, as well as the latest label (earlier labels are not used to predict future labels from the same patient, following OHDSI’s cohort-defining guidelines \citep{ohdsi-book-2019}).

\subsection{Problem setup} 

We consider supervised prediction from longitudinal EHR data. For each patient $i\in\{1,\ldots,N\}$ in a dataset of $N$ total patients, we fix a prediction time $\tau_i$ and collect all structured events prior to $\tau_i$. $S_i=\{(c_{ij}, t_{ij})\}_{j=1}^{T_i}$ therefore represents the event stream of patient $i$ \citep{mcdermott2023event}, where $c_{ij}\in\mathcal{C}$ are OMOP concept IDs and $t_{ij}<\tau_i$ are the corresponding time-stamps of these events. The binary task-specific label is $y_i\in\{1,\dots,K\} \in \{0,1\}$.

\subsection{Count-based models}

To construct tabular representations, we apply a map $m:\mathcal{C}\!\to\!\mathcal{G}$ to aggregate fine-grained concepts into ICD-10/ATC section categories. For conditions, concept identifiers were first mapped to their corresponding ICD-10 codes and then rolled up into higher-level diagnostic categories (\textit{e.g.}, ``I20-I25: Ischemic heart diseases”). For drug exposures, drug identifiers were first mapped to ATC codes and subsequently aggregated at the pharmacological section level (\textit{e.g.}, ``C07: Beta blocking agents”). This ontology-based roll-up strategy reduces feature sparsity and improves performance, as has been shown in prior EHR-based data competitions \citep{bergquist2023evaluation, bergquist2024crowd}. 

We define two look-back windows relative to $\tau_i$: $\mathcal{W}_{\mathrm{recent}}=[\tau_i-365\text{ days},\tau_i)$ and $\mathcal{W}_{\mathrm{history}}=(-\infty,\tau_i-365\text{ days})$. This time-stratified approach allows us to capture both the recent, as well as the long-term history of diagnoses and medication exposure. 

For category $g\in\mathcal{G}$ and window $w\in\{\mathrm{recent},\mathrm{history}\}$, the count-based feature is
\[
n^{(w)}_{ig} \;=\; \sum_{j=1}^{T_i} \mathbf{1}\!\left[m(c_{ij})=g \;\wedge\; t_{ij}\in \mathcal{W}_w\right].
\]
The count-based feature representation for patient $i$ is 
\[
\mathbf{x}_{i,count} \;=\; \big[\,n^{(\mathrm{recent})}_{ig},\, n^{(\mathrm{history})}_{ig}\,\big]_{g\in\mathcal{G}}
\;\in\; \mathbb{R}^{2|\mathcal{G}|}
\]

In our experiment, $|\mathcal{G}| = \text{Total number of ICD-10 and ATC sections} = 388$.

Given the count-based feature representation $\mathbf{x}_{i,count}$ for each patient $i$, we pass them to two representative tabular models: LightGBM \citep{ke2017lightgbm} and TabPFN \citep{hollmann2022tabpfn}. LightGBM is a gradient boosting framework that builds an ensemble of shallow decision trees, where each tree focuses on correcting the mistakes of the previous ones. It is widely used in structured data competitions and clinical prediction tasks, where it achieved outstanding performance \citep{bergquist2023evaluation, bergquist2024crowd}. TabPFN is a recently developed foundation model for tabular data. Instead of training a new model for each dataset, TabPFN has been pretrained on millions of synthetic classification problems and produces a learned Bayes predictive distribution in a single forward pass conditioned on the entire training data (features and labels) and the test features, yielding the posterior predictive label distribution without task-specific retraining. TabPFN has been reported to outperform boosting models on small- to medium-sized tabular datasets \citep{hollmann2022tabpfn}.

\subsection{Pretrained transformers}

We also evaluated CLMBR (Clinical Language Model-Based Representations) \citep{steinberg2021language, guo2024multi}, a pretrained transformer model designed for structured EHR data and introduced together with the EHRSHOT dataset. Patient histories were organized as longitudinal sequences of timestamped medical events, with each medical concept (\textit{e.g.}, diagnoses and medications) represented as a token. CLMBR learns contextual embeddings of these sequences using a masked language modeling objective, allowing the model to capture temporal dependencies and co-occurrence patterns among medical concepts. The learned patient representations can then be used as input to lightweight classifiers for downstream prediction tasks. 

Formally, given the original event stream $S_i=\{(c_{ij},t_{ij})\}_{j=1}^{T_i}$ with $c_{ij}\in\mathcal{C}$ (no roll-up), the CLMBR yields contextual states $\{\mathbf{h}_{ij}\}$; CLMBR uses a transformer $f_{\theta}$ to form a fixed-length patient representation
\[
\mathbf{r}_i \;=\; f_{\theta}(S_i).
\]

This vector $\mathbf{r}_i$ is then passed to a downstream classifier.
We used LightGBM and TabPFN as classifier, trained on $\{(\mathbf{r}_i,y_i)\}$ in the training set, and evaluated on the test set.

\subsection{Mixture-of-agents pipelines}

The MoA pipeline adapts LLMs to structured EHR by introducing a collaborative multi-agent workflow. Let $M_i$ denote the plain-text serialization of $S_i$, formatted as a time-ordered sequence of \emph{(medical concept name, age at event)} pairs. In this setting, a summarizer agent $\mathcal{A}$ first converts $M_i$ into readable and concise textual summaries. 
\[
U_i \;=\; \mathcal{A}(M_i),
\]
which is then consumed by a predictor $\mathcal{P}$ to produce class predictions per patient:
\[
\hat{p}_i \;=\; \mathcal{P}\!\big(U_i\big).
\]
In our implementation, we adopted Llama-3-8b \citep{llama3modelcard}, reported to perform well with structured EHR data \citep{chen2024clinicalbench, gao2025moma}, and Qwen2.5-14B-Instruct, which has demonstrated stronger reasoning ability in long-context settings \citep{qwen2.5}. Both models were prompted to summarize raw tabular data into clinically meaningful structured text (prompts in Supplementary Section \ref{prompt:long_los} to \autoref{prompt:lupus}), and the resulting summaries served as input for the downstream classifiers. We used BGE-large-en-v1.5 \citep{bge_embedding} and ClinicalBERT \citep{alsentzer2019publicly} as classifiers due to their outstanding performance on clinical text representation tasks \citep{myers2025evaluating, gao2024automated}. In particular, we extracted the final hidden state corresponding to the [CLS] token and passed it through a single feed-forward layer, yielding the logit predictions for classification.

\section{Results}

In this section, we report metrics for the assessed methods across eight tasks, highlight the MoA's intermediate summary quality, and examine medical concepts contributing to prediction.

\subsection{Discrimination performance}

\begin{figure}[ht!]
    \centering
    \includegraphics[width=1.00\textwidth]{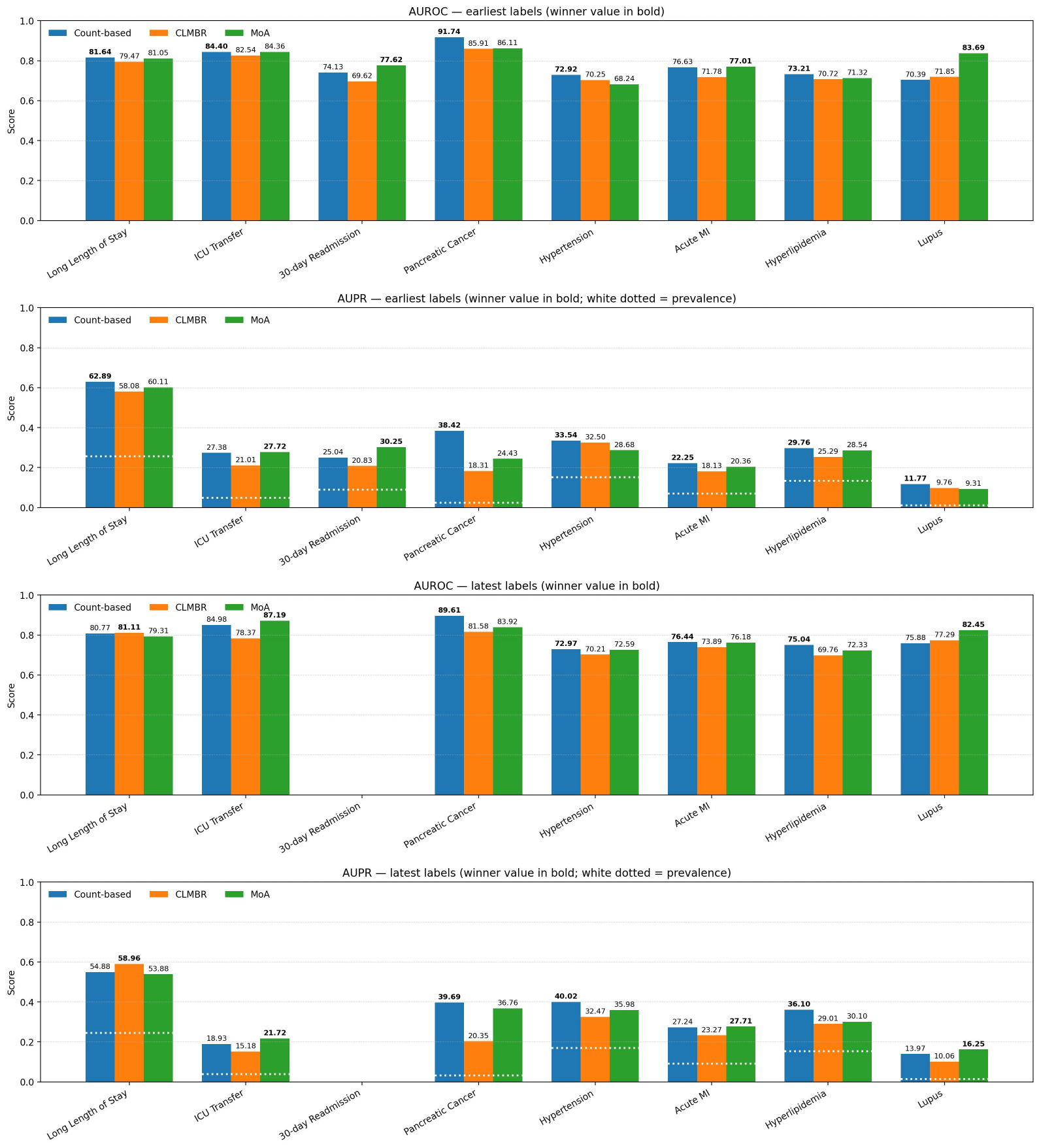}
    \caption{\textbf{Performance of the benchmarked models across the eight prediction tasks in EHRSHOT under two label definitions (earliest and latest).} Top panels show AUROC; bottom panels show AUPR. Bolded bars indicate the best-performing model for each task; the white dotted line in each AUPR panel marks the outcome prevalence. Across tasks, wins are shared mainly between the count-based methods and the MoA pipeline, with count-based methods holding a slight overall edge and generally outperforming CLMBR.}
    \label{fig:main_results}
\end{figure}

Figure \ref{fig:main_results} shows the results of our evaluations, and \ref{tab:main_results_earliest} and \ref{tab:main_results_latest} report the corresponding numerical results. Across tasks, wins concentrate in the count-based and MoA methodologies. For clarity, Figure \ref{fig:main_results} only reports a single winning method within each of the three methodology categories: (i) count-based features with LightGBM \textit{vs.} count-based features with TabPFN, (ii) CLMBR embeddings with LightGBM \textit{vs.} CLMBR embeddings with TabPFN, and (iii) the best LLM choice within the MoA pipeline. The results of other variants for each methodology category are shown in \ref{tab:main_results_all_AUROC_earliest} to \ref{tab:count_latest_aupr}. In the earliest-label setting, count-based methods take 5/8 AUROC wins (MoA: 3/8) and 6/8 AUPR wins (MoA: 2/8). In the latest-label setting and excluding readmission here due to its abnormally low prevalence (see \ref{tab:dataset}), count-based methods lead AUROC with 4/7 wins (MoA: 2/7; CLMBR: 1/7) and split AUPR (3/7) with MoA (3/7) and CLMBR (1/7). In short, most wins are shared by count-based and MoA methods, with count-based finishing slightly ahead overall and consistently outperforming CLMBR. Therefore, count-based methods remain a strong baseline for structured EHR benchmarking, while MoA methods offer gains on specific tasks.

\subsection{Quality of the intermediate summary}

\begin{figure}[ht!]
    \centering
    \includegraphics[width=1.00\textwidth]{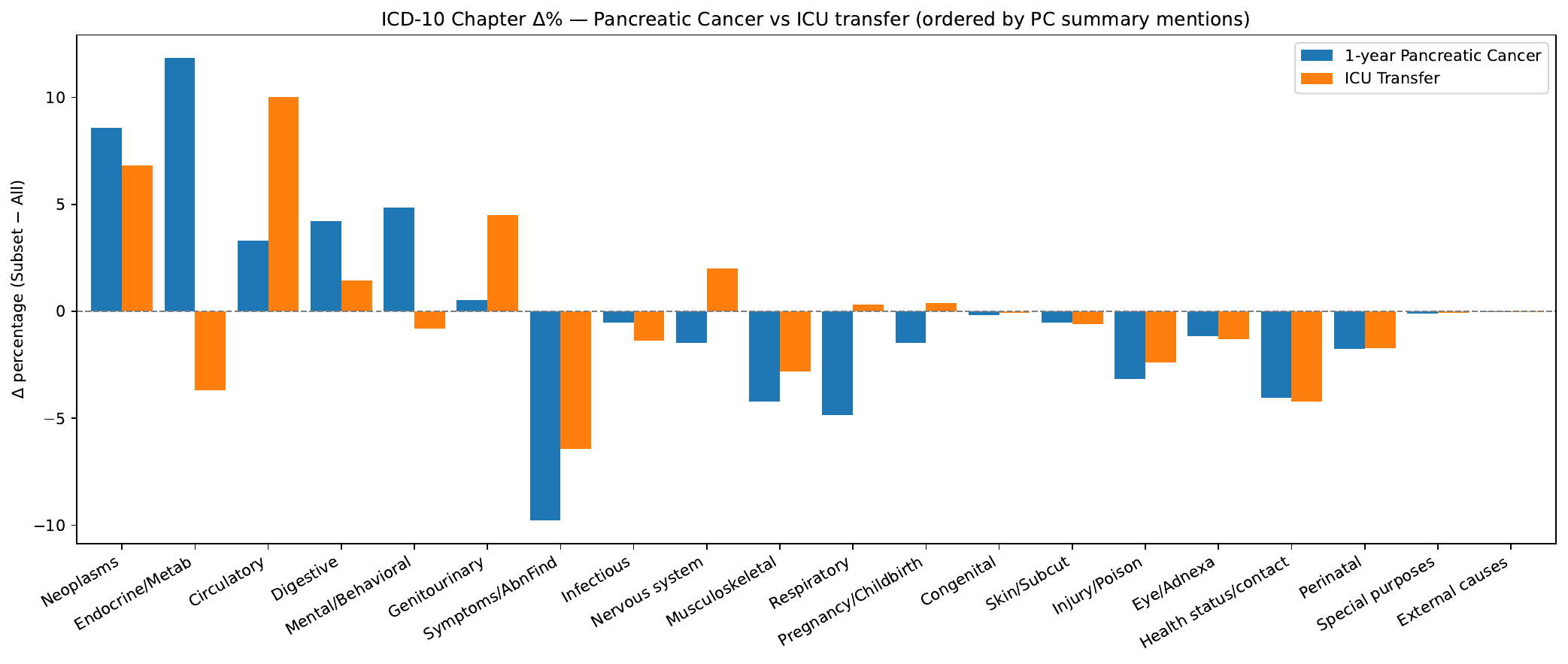}
    \caption{\textbf{Task-adaptive emphasis in MoA intermediate summaries.} 
    Bars show the change in ICD-10 chapter mention percentages between the MoA summary and the original structured EHR (\(\Delta\%\!=\) percentage mentions in summary \( - \) percentage mentions in original), computed 
    with MedCat concept extraction. Positive values indicate chapters the summary
    amplifies; negative values indicate down-weighting. Chapters with opposite sign directions 
    illustrate adaptation to the prediction task. For example, \textit{Endocrine/Metabolic} is amplified 
    for pancreatic cancer but reduced for ICU transfer, whereas \textit{Nervous system} is boosted for ICU transfer and de-emphasized for pancreatic cancer.}
    \label{fig:chapter_diff}
\end{figure}

In the MoA pipeline, an LLM agent is prompted to produce intermediate summaries relevant to the prediction target. To compare the differences in information content between the original text (raw structured EHR) and the intermediate summary, we first used MedCat \citep{kraljevic2021multi} to extract concepts as ICD-10 chapters from both texts, computed the percentages of mentions of each ICD-10 chapter, and plotted the percentage change (summary minus original EHR) in Figure \ref{fig:chapter_diff}. Positive values indicate that the summary amplifies the chapter, while negative values indicate that information from the chapter is down-weighted. Chapters which move in opposite directions depending on the task highlight how the MoA pipeline adapts to the clinical prediction target. For example, for pancreatic cancer prediction, the summary amplifies Endocrine/Metabolic and Mental/Behavioral content, while down-weighting it for ICU transfer prediction. Conversely, the ICU-transfer summary elevates Nervous system mentions, while de-emphasizing them in the pancreatic cancer summary. This assessment directly shows that intermediate summaries help steer the input information content towards the prediction task.

\subsection{A case study for the Mixture-of-agent pipeline}

\begin{figure}[ht!]
    \centering
    \includegraphics[width=0.90\textwidth]{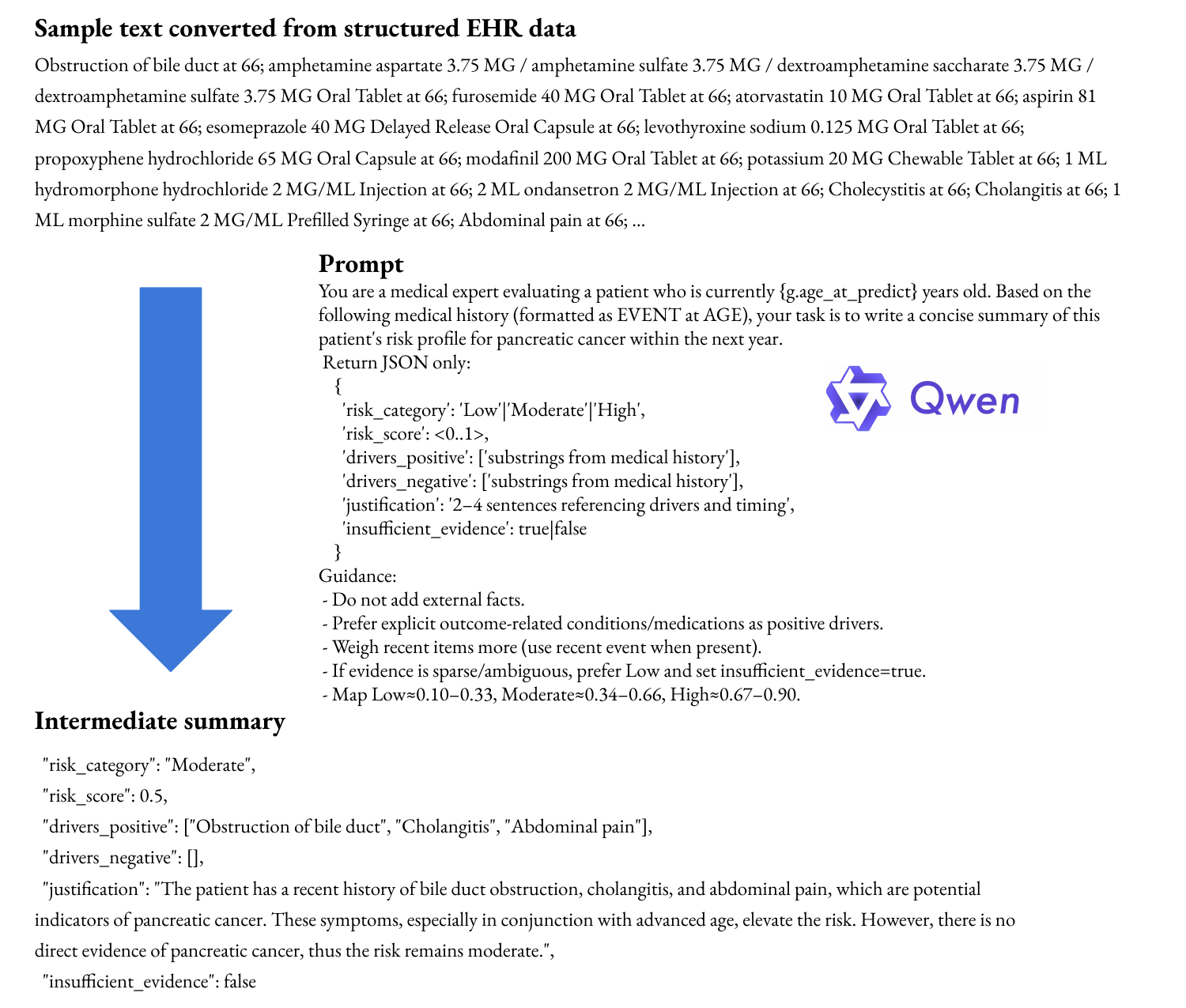}
    \caption{\textbf{Case study: LLM-generated intermediate summary from structured EHR in the MoA pipeline.} Structured EHR is first converted into an ``EVENT at AGE'' text sequence and further passed to Qwen with a task prompt to predict the risk of a new pancreatic cancer diagnosis in the next year. The model is instructed to return a constrained JSON object containing \texttt{risk\_category} (Low/Moderate/High), \texttt{risk\_score} \([0,1]\), \texttt{drivers\_positive}/\texttt{drivers\_negative} from the input, a 2 to 4 sentence \texttt{justification} that references evidence and timing, and an \texttt{insufficient\_evidence} flag. In this example, Qwen highlights biliary obstruction, cholangitis, and abdominal pain, and returns a \emph{Moderate} risk with score 0.5, while also yielding an interpretable intermediate summary used for downstream text classifiers in the MoA pipeline.}
        \label{fig:demo_moa}
\end{figure}

The case study in Figure \ref{fig:demo_moa} shows how the MoA pipeline summarizes a patient’s recent history for pancreatic cancer prediction. The structured EHR data are first converted into “EVENT at AGE” snippets and passed to the summarizer (Qwen2.5-14B-Instruct). The prompt requires a JSON output with a risk category/score, positive and negative drivers, as well as a short justification. In this example, given the input EHR text (\textit{e.g.}, “obstruction of bile duct,” “cholangitis,” “abdominal pain", "multiple analgesics"), Qwen produces an intermediate summary that labels the case as Moderate risk with a score of 0.5, cites biliary obstruction and cholangitis as positive drivers, and briefly explains its underlying rationale. When ingested by the downstream text classifier, this evidence yields towards a positive prediction, which is consistent with the true positive label for this patient.

\subsection{SHAP values of LightGBM}

\begin{figure}[ht!]
    \centering
    \includegraphics[width=\textwidth]{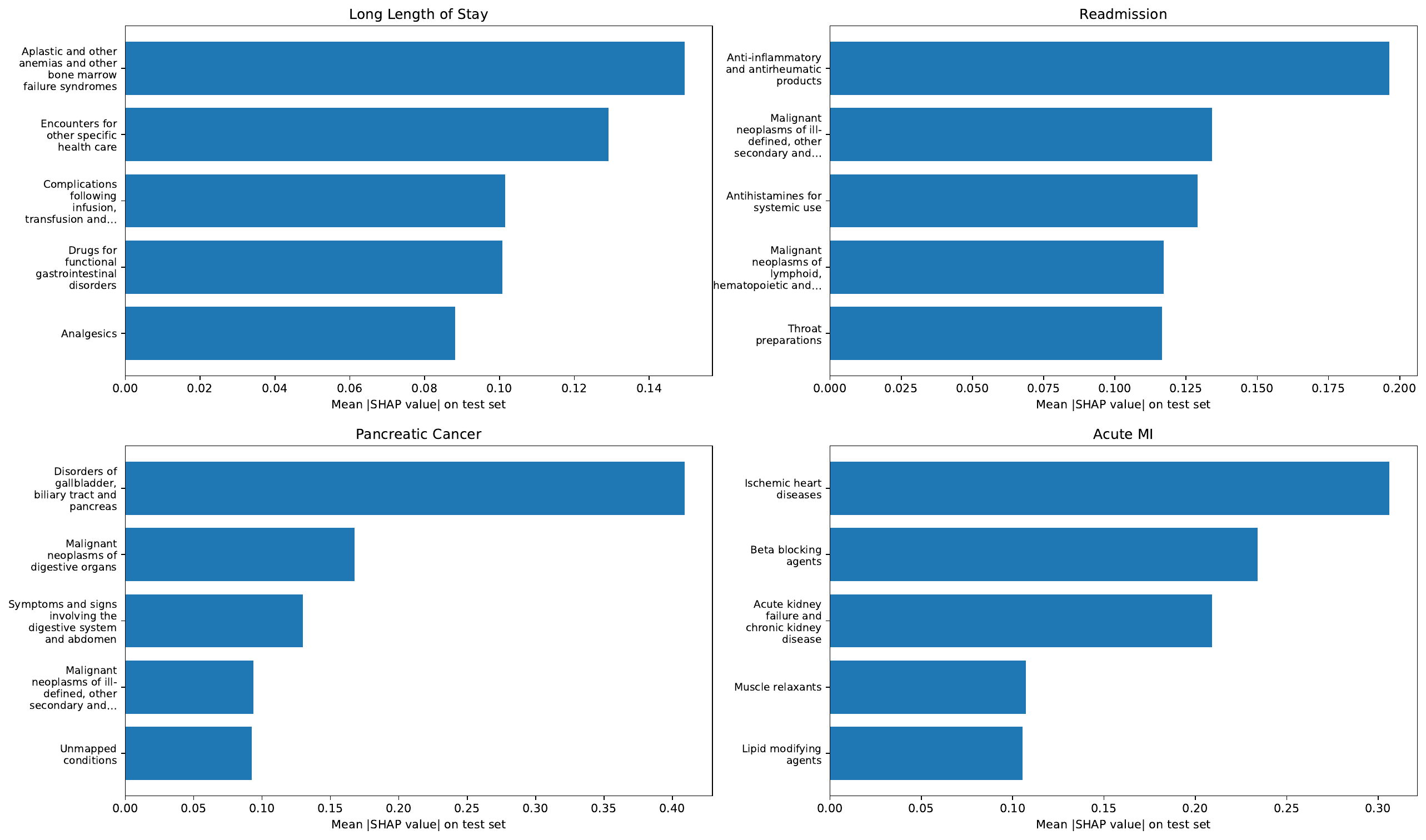}
    \caption{\textbf{SHAP analysis of LightGBM models for four prediction tasks}. For each task, we show the mean absolute SHAP values on the test set for the top five most influential features. All of the contributing factors come from the most recent time bin (within one year before the prediction time). This detail is not explicitly labeled in the figure, as the concept names are already lengthy.}
    \label{fig:shap}
\end{figure}

To better understand the risk factors driving model predictions, we examined the SHAP values \citep{lundberg2017unified} from the LightGBM model features (Figure \ref{fig:shap}). All the top five contributing factors came from the most recent time bin (within one year before prediction time), suggesting that events in the immediate history carry the strongest signal for prediction, while factors from longer look-back windows may not have a substantial influence on the outcomes.

For long length-of-stay, hematologic disorders such as aplastic anemia, complications following transfusion, and frequent health care encounters were among the strongest predictors, consistent with patients requiring prolonged and complex hospital care. In the readmission task, medication use and cancer-related diagnoses showed high importance. The model highlighted anti-inflammatory drugs, systemic antihistamines, and throat preparations, along with malignant neoplasms, suggesting that both chronic disease burden and frequent medication use contribute to higher readmission risk. For pancreatic cancer prediction, diagnoses related to hepatobiliary and digestive disorders were most influential, which is expected given the biological links between gallbladder, biliary tract, and pancreatic diseases. Symptoms such as abdominal complaints and general digestive neoplasms also appeared as strong predictors. In acute myocardial infarction, cardiovascular conditions were the most prominent. Ischemic heart disease and beta-blocker prescriptions were the leading contributing factors, along with kidney disease, lipid-lowering therapy, and muscle relaxants, reflecting established risk factors and common co-treatments in patients at high risk of acute cardiac events.

\section{Discussion}

In this study, we benchmarked three categories of methods for clinical prediction using structured EHR data: (i) count-based models using ontology roll-ups and time-binned features, (ii) a pretrained sequential transformer (CLMBR), and (iii) a mixture-of-agents LLM pipeline that summarizes tabular histories before classification. We evaluated eight binary prediction tasks on the EHRSHOT dataset: Long Length of Stay, ICU Transfer, Readmission, Pancreatic Cancer, Hypertension, Acute Myocardial Infarction, and Hyperlipidemia. Across tasks, victories are split primarily between the count-based models and the MoA pipeline, with count-based methods holding a slight overall edge and generally outperforming CLMBR. This shows how, despite recent advances in multi-agent LLM pipelines, count-based approaches remain a strong, practical choice for structured EHR prediction.

To our knowledge, this is the first benchmark to compare count-based models head-to-head with a MoA pipeline on structured EHR data. Although MoA pipelines have demonstrated superior performance compared to single LLM pipelines, previous studies have largely only evaluated single LLM baselines on structured EHR data \citep{chen2024clinicalbench}. To fill in this knowledge gap, we hereby adapted and evaluated a MoA pipeline designed specifically for tabular data. We also equipped the count-based approach with an up-to-date tabular foundation model (TabPFN) in addition to a strong GBM baseline (LightGBM), offering a fair and contemporary comparison against sequential and LLM-based methods under the same cohorts and splits.

Although predicting the latest label for a patient provides richer contextual information compared to using the earliest label (\ref{tab:dataset}), overall performance patterns remain largely unchanged. In the latest label setting, count-based models and the MoA pipeline still dominated the evaluation tasks. The only exception is that in the Long Length of Stay task with the latest labels, CLMBR shows an outstanding improvement and takes the lead. This suggests that count-based models and the MoA pipeline are comparatively robust to context richness, whereas the transformer-based CLMBR may benefit more from more contextual details.

Our interpretability analysis for the count-based methods shows that the top five SHAP contributors come from the most recent time bin (within one year before prediction time). This suggests that immediate clinical history dominates model decisions, while long-past events contribute with less signal. One implication of these findings is that fine-grained temporal ordering and long-range dependencies may be less impactful for the prediction tasks assessed here in the EHRSHOT dataset.

We performed a series of ablation studies on the MoA pipelines and the count-based baselines. For the MoA pipeline, we varied both the summarizing agent (Llama-3-8b-instruct \textit{vs.} Qwen2.5-14B-Instruct) as well as the classifier (ClinicalBERT \textit{vs.} BGE-large-en-v1.5), and we also evaluated prompts that produce unstructured free-text summaries (Supplementary Section \ref{prompt:unstructured_long_los} to \ref{prompt:unstructured_lupus}) rather than the constrained JSON. As reported in \ref{tab:moa_earliest_auroc} to \ref{tab:count_latest_aupr}, the combination Qwen2.5-14B-Instruct + BGE-large-en-v1.5 with structured prompts yielded the strongest performance among all MoA variants, underscoring that agent choice and prompt format substantially affect the performance of MoA pipelines. For the count-based models, we compared pipelines with \textit{vs.} without ontology roll-up; in every task, ontology roll-up improved performance over the non-rolled counterparts.

Several limitations warrant mention. Our analysis is based on a single-institution dataset and a limited set of outcomes, which constrains external validity. We also focused only on diagnoses and medications training data. Further integration of labs, vitals, and procedures in model training might improve performance. Future work should expand across multiple institutions, include a broader set of clinical information resources, and test additional LLM agents and summarization strategies.

Overall, our results reaffirm the strength of count-based modeling for structured EHR prediction, even in the era of LLMs. With up-to-date LLM agents, the MoA pipeline achieves superior performance in a subset of the evaluation tasks. These findings highlight that traditional methods still provide strong baselines, but generative and transformer-based approaches open new possibilities under specific clinical scenarios. As clinical AI moves toward foundation-model paradigms \citep{moor2023foundation, wornow2023shaky}, it will be worthwhile to systematically evaluate how these approaches leverage richer histories, scale across diverse health systems, and integrate into decision-making workflows.

\clearpage
\bibliographystyle{plainnat}
\bibliography{reference.bib}

\clearpage
\setcounter{page}{1}
\appendix
\section*{Supplementary Material}

\setcounter{table}{0}
\renewcommand{\thetable}{Supplementary Table \arabic{table}} 

\ref{tab:dataset} summarizes the cohorts across the eight prediction tasks.

\begin{table}[!htbp]
\centering
\caption{Cohort characteristic}
\adjustbox{scale=1.10,center}{
\resizebox{\textwidth}{!}{
\begin{tabular}{*5l}
\toprule
{\textbf{Task}}   &  Long LOS &  Readmission &  Pancreatic Cancer &  Acute MI \\
\midrule
\textbf{Cohort characteristics} & & & & \\
Encounters, n & 3,855 & 3,718 & 3,864 & 3,834\\ 
Age, median (IQR) & 58 (40, 69) & 58 (43, 69) & 58 (41, 69) & 58 (41, 69)\\
Female, n (\%) & 2,067 (53.6) & 1,977 (53.2) & 2,071 (53.6) & 2,071 (54.0)\\
White, n (\%) & 2,248 (58.3) & 2,190 (58.9) & 2,251 (58.3) & 2,231 (58.2)\\
\midrule
\textbf{Outcome and context information: earliest (if multiple)} & & & & \\
Prevalence (\%) & 24.9 & 8.2 & 4.6 & 7.1 \\
Number of distinct diagnoses, median (IQR) & 19 (10, 34) & 21 (11, 38) & 21 (11, 37) & 21 (11, 36)  \\
Number of diagnoses, median (IQR) & 31 (15, 67) & 44 (18, 99) & 43 (18, 97) & 43 (18, 95.5)\\
Number of distinct medications, median (IQR) & 21 (10, 35) & 37 (22, 56)  & 36 (21.5, 55) & 36 (21, 55)  \\
Number of medications, median (IQR) & 25 (11, 43) & 75 (40, 150)  & 73 (38, 148) & 73 (38, 146)  \\
Number of distinct dates, median (IQR) & 9 (4, 24) & 16 (8, 32) & 15 (8, 31) & 15 (8, 31)\\
\midrule
\textbf{Outcome and context information: most recent (if multiple)} & & & & \\
Prevalence (\%) & 24.3 & 0.5 & 5.5 & 9.3 \\
Number of distinct diagnoses, median (IQR) & 28 (14, 55) & 32 (15, 60)  & 30 (15, 57) & 29.5 (15, 55)  \\
Number of diagnoses, median (IQR) & 21 (55, 163) & 77 (29, 211)  & 73 (26, 195) & 70 (26, 184.75)  \\
Number of distinct medications, median (IQR) & 33 (15, 63) & 50 (30, 81) & 36 (21.5, 55) & 36 (21, 55) \\
Number of medications, median (IQR) & 43 (17, 167) & 121 (55, 306) & 73 (38, 148) & 73 (38, 146) \\
Number of distinct dates, median (IQR) & 19 (6, 58) & 27 (11, 69)  & 23 (10, 59.75) & 23 (10, 57)  \\
\midrule
& & & & \\
\midrule
\textbf{Task}   &  ICU transfer &  Hypertension &  Hyperlipidemia &  Lupus \\
\midrule
\textbf{Cohort characteristics} & & & & \\
Encounters, n & 3,617 & 2,328 & 2,650 & 3,864\\ 
Age, median (IQR) & 58 (41, 69) & 50 (35, 64) & 51 (35, 65) & 58 (41, 69)  \\
Female, n (\%) &  1,959 (54.2) & 1,416 (60.8) & 1,594 (60.2) & 2,058 (53.3)  \\
White, n (\%) & 2,103 (58.1) & 1,355 (58.2) & 1,529 (57.7) & 2,253 (58.3)  \\
\midrule
\textbf{Outcome and context information: earliest (if multiple)} & & & & \\
Prevalence (\%) & 4.5 & 14.3 & 12.7 & 2.6  \\
Number of distinct diagnoses, median (IQR) & 19 (10, 34) & 17 (9, 31) & 17 (9, 31) & 21 (11, 37)  \\
Number of diagnoses, median (IQR) & 31 (15, 73) & 32 (13, 76) & 34 (14, 77) & 43 (18, 96)\\
Number of distinct medications, median (IQR) & 22 (11, 36) & 30 (17, 46.25) & 31 (18, 49) & 36 (22, 55)  \\
Number of medications, median (IQR) & 25 (11, 46.75) & 58 (30, 118) & 61 (32, 126) & 73 (38, 147) \\
Number of distinct dates, median (IQR) & 10 (4, 26) & 13 (7, 27) & 13 (7, 27) & 15 (8, 31)\\
\midrule
\textbf{Outcome and context information: most recent (if multiple)} & & & & \\
Prevalence (\%) & 4.6 & 16.5 & 15.4 & 3.2 \\
Number of distinct diagnoses, median (IQR) & 28 (14, 55) & 22 (10, 41.5) & 23 (11, 45) & 30 (15, 57)  \\
Number of diagnoses, median (IQR) & 56 (20, 169) & 46 (17, 123) & 51 (19, 138) & 73 (26, 195)\\
Number of distinct medications, median (IQR) & 33 (15, 64) & 37 (20, 60) & 40 (22, 66) & 48 (28, 78)  \\
Number of medications, median (IQR) & 44 (17, 173) & 76 (36, 194) & 86.5 (39, 226.25) & 113 (51, 283.75) \\
Number of distinct dates, median (IQR) & 20 (6, 59) & 18 (8, 43) & 19 (9, 46) & 25 (10, 63)\\
\bottomrule
\end{tabular}
}}
\label{tab:dataset}
\end{table}

\ref{tab:main_results_earliest} shows the numerical benchmark results using the earliest visit as the prediction time.

\begin{table}[ht]
\centering
\caption{Earliest labels — best of each group (row-wise winners in \textbf{bold}).}
\setlength{\tabcolsep}{6pt}
\label{tab:main_results_earliest}
\begin{tabular}{lcccccc}
\toprule
& \multicolumn{3}{c}{AUROC} & \multicolumn{3}{c}{AUPR} \\
\cmidrule(lr){2-4}\cmidrule(lr){5-7}
Task & Count-based & CLMBR & MoA & Count-based & CLMBR & MoA \\
\midrule
Long LOS            & \textbf{0.8164} & 0.7947 & 0.8105 & \textbf{0.6289} & 0.5808 & 0.6011 \\
ICU Transfer             & \textbf{0.8440} & 0.8254 & 0.8436 & 0.2738 & 0.2101 & \textbf{0.2772} \\
Readmission    & 0.7413 & 0.6962 & \textbf{0.7762} & 0.2504 & 0.2083 & \textbf{0.3025} \\
Pancreatic Cancer        & \textbf{0.9174} & 0.8591 & 0.8611 & \textbf{0.3842} & 0.1831 & 0.2443 \\
Hypertension   & \textbf{0.7292} & 0.7025 & 0.6824 & \textbf{0.3354} & 0.3250 & 0.2868 \\
Acute MI       & 0.7663 & 0.7178 & \textbf{0.7701} & \textbf{0.2225} & 0.1813 & 0.2036 \\
Hyperlipidemia & \textbf{0.7321} & 0.7072 & 0.7132 & \textbf{0.2976} & 0.2529 & 0.2854 \\
Lupus          & 0.7039 & 0.7185 & \textbf{0.8369} & \textbf{0.1177} & 0.0976 & 0.0931 \\
\bottomrule
\end{tabular}
\end{table}

\ref{tab:main_results_latest} shows the numerical benchmark results using the last visit as the prediction time.

\begin{table}[ht]
\centering
\caption{Latest labels — best of each group (row-wise winners in \textbf{bold}).}
\setlength{\tabcolsep}{6pt}
\label{tab:main_results_latest}
\begin{tabular}{lcccccc}
\toprule
& \multicolumn{3}{c}{AUROC} & \multicolumn{3}{c}{AUPR} \\
\cmidrule(lr){2-4}\cmidrule(lr){5-7}
Task & Count-based & CLMBR & MoA & Count-based & CLMBR & MoA \\
\midrule
Long LOS            & 0.8077 & \textbf{0.8111} & 0.7931 & 0.5488 & \textbf{0.5896} & 0.5388 \\
ICU Transfer           & 0.8498 & 0.7837 & \textbf{0.8719} & 0.1893 & 0.1518 & \textbf{0.2172} \\
Readmission    & N/A & N/A & N/A & N/A & N/A & N/A \\
Pancreatic Cancer       & \textbf{0.8961} & 0.8158 & 0.8392 & \textbf{0.3969} & 0.2035 & 0.3676 \\
Hypertension   & \textbf{0.7297} & 0.7021 & 0.7259 & \textbf{0.4002} & 0.3247 & 0.3598 \\
Acute MI        & \textbf{0.7644} & 0.7389 & 0.7618 & 0.2724 & 0.2327 & \textbf{0.2771} \\
Hyperlipidemia & \textbf{0.7504} & 0.6976 & 0.7233 & \textbf{0.3610} & 0.2901 & 0.3010 \\
Lupus          & 0.7588 & 0.7729 & \textbf{0.8245} & 0.1397 & 0.1006 & \textbf{0.1625} \\
\bottomrule
\end{tabular}
\end{table}

\ref{tab:main_results_all_AUROC_earliest}, \ref{tab:main_results_all_AUPR_earliest}, \ref{tab:main_results_all_AUROC_latest}, and \ref{tab:main_results_all_AUPR_latest} show results of count-based and CLMBR variants.

\begin{table}[ht]
\centering
\caption{Earliest labels — \textbf{AUROC} (mean $\pm$ sd) across tasks and methods for count-based and CLMBR variants.}
\label{tab:main_results_all_AUROC_earliest}
\scriptsize
\setlength{\tabcolsep}{5pt}
\begin{tabular}{lccccc}
\toprule
\textbf{Task} & \textbf{Count-based + LGBM} & \textbf{Count-based + TabPFN} & \textbf{CLMBR + LGBM} & \textbf{CLMBR + TabPFN} & \textbf{MoA} \\
\midrule
Long LOS          & $0.8047 \pm 0.0039$ & \textbf{\boldmath$0.8164 \pm \text{N/A}$} & $0.7835 \pm 0.0072$ & $0.7947 \pm \text{N/A}$ & $0.8105 \pm 0.0257$ \\
ICU Transfer      & \textbf{\boldmath$0.8440 \pm 0.0058$} & $0.8377 \pm \text{N/A}$ & $0.7456 \pm 0.0156$ & $0.8254 \pm \text{N/A}$ & $0.8436 \pm 0.0054$ \\
Readmission       & $0.7413 \pm 0.0032$ & $0.7366 \pm \text{N/A}$ & $0.6962 \pm 0.0093$ & $0.6320 \pm \text{N/A}$ & \textbf{\boldmath$0.7762 \pm 0.0176$} \\
Pancreatic Cancer & $0.8821 \pm 0.0070$ & \textbf{\boldmath$0.9174 \pm \text{N/A}$} & $0.8004 \pm 0.0087$ & $0.8591 \pm \text{N/A}$ & $0.8611 \pm 0.0150$ \\
Hypertension      & $0.7233 \pm 0.0041$ & \textbf{\boldmath$0.7292 \pm \text{N/A}$} & $0.6956 \pm 0.0105$ & $0.7025 \pm \text{N/A}$ & $0.6824 \pm 0.0256$ \\
Acute MI          & $0.7663 \pm 0.0119$ & $0.7590 \pm \text{N/A}$ & $0.7093 \pm 0.0161$ & $0.7178 \pm \text{N/A}$ & \textbf{\boldmath$0.7701 \pm 0.0091$} \\
Hyperlipidemia    & $0.7289 \pm 0.0044$ & \textbf{\boldmath$0.7321 \pm \text{N/A}$} & $0.6592 \pm 0.0041$ & $0.7072 \pm \text{N/A}$ & $0.7132 \pm 0.0140$ \\
Lupus             & $0.6689 \pm 0.0100$ & $0.7039 \pm \text{N/A}$ & $0.7079 \pm 0.0107$ & $0.7185 \pm \text{N/A}$ & \textbf{\boldmath$0.8369 \pm 0.0129$} \\
\bottomrule
\end{tabular}
\end{table}

\begin{table}[ht]
\centering
\caption{Earliest labels — \textbf{AUPR} (mean $\pm$ sd) across tasks and methods for count-based and CLMBR variants..}
\label{tab:main_results_all_AUPR_earliest}
\scriptsize
\setlength{\tabcolsep}{5pt}
\begin{tabular}{lccccc}
\toprule
\textbf{Task} & \textbf{Count-based + LGBM} & \textbf{Count-based + TabPFN} & \textbf{CLMBR + LGBM} & \textbf{CLMBR + TabPFN} & \textbf{MoA} \\
\midrule
Long LOS          & $0.6172 \pm 0.0116$ & \textbf{\boldmath$0.6289 \pm \text{N/A}$} & $0.5673 \pm 0.0137$ & $0.5808 \pm \text{N/A}$ & $0.6011 \pm 0.0533$ \\
ICU Transfer      & $0.2381 \pm 0.0220$ & $0.2738 \pm \text{N/A}$ & $0.1519 \pm 0.0206$ & $0.2101 \pm \text{N/A}$ & \textbf{\boldmath$0.2772 \pm 0.0089$} \\
Readmission       & $0.2504 \pm 0.0203$ & $0.2449 \pm \text{N/A}$ & $0.2083 \pm 0.0027$ & $0.1603 \pm \text{N/A}$ & \textbf{\boldmath$0.3025 \pm 0.0225$} \\
Pancreatic Cancer & $0.3144 \pm 0.0463$ & \textbf{$0.3842 \pm \text{N/A}$} & $0.1722 \pm 0.0313$ & $0.1831 \pm \text{N/A}$ & $0.2443 \pm 0.0636$ \\
Hypertension      & $0.3241 \pm 0.0070$ & \textbf{\boldmath$0.3354 \pm \text{N/A}$} & $0.3188 \pm 0.0064$ & $0.3250 \pm \text{N/A}$ & $0.2868 \pm 0.0327$ \\
Acute MI          & $0.2113 \pm 0.0052$ & \textbf{\boldmath$0.2225 \pm \text{N/A}$} & $0.1735 \pm 0.0045$ & $0.1813 \pm \text{N/A}$ & $0.2036 \pm 0.0169$ \\
Hyperlipidemia    & \textbf{\boldmath$0.2976 \pm 0.0124$} & $0.2955 \pm \text{N/A}$ & $0.2029 \pm 0.0077$ & $0.2529 \pm \text{N/A}$ & $0.2854 \pm 0.0231$ \\
Lupus             & $0.0866 \pm 0.0252$ & \textbf{\boldmath$0.1177 \pm \text{N/A}$} & $0.0976 \pm 0.0469$ & $0.0608 \pm \text{N/A}$ & $0.0931 \pm 0.0418$ \\
\bottomrule
\end{tabular}
\end{table}

\begin{table}[ht]
\centering
\caption{Latest labels — \textbf{AUROC} (mean $\pm$ sd) for count-based and CLMBR variants.}
\label{tab:main_results_all_AUROC_latest}
\scriptsize
\setlength{\tabcolsep}{5pt}
\begin{tabular}{lccccc}
\toprule
\textbf{Task} & \textbf{Count-based + LGBM} & \textbf{Count-based + TabPFN} & \textbf{CLMBR + LGBM} & \textbf{CLMBR + TabPFN} & \textbf{MoA} \\
\midrule
Long LOS           & $0.7918 \pm 0.0015$ & $0.8077 \pm \text{N/A}$ & $0.7851 \pm 0.0037$ & \textbf{\boldmath$0.8111 \pm \text{N/A}$} & $0.7931 \pm 0.0030$ \\
ICU Transfer       & $0.8213 \pm 0.0036$ & $0.8498 \pm \text{N/A}$ & $0.7837 \pm 0.0090$ & $0.7680 \pm \text{N/A}$ & \textbf{\boldmath$0.8719 \pm 0.0169$} \\
Readmission        & $\text{N/A} \pm \text{N/A}$ & $\text{N/A} \pm \text{N/A}$ & $\text{N/A} \pm \text{N/A}$ & $\text{N/A} \pm \text{N/A}$ & $\text{N/A} \pm \text{N/A}$ \\
Pancreatic Cancer  & $0.8609 \pm 0.0113$ & \textbf{\boldmath$0.8961 \pm \text{N/A}$} & $0.7867 \pm 0.0039$ & $0.8158 \pm \text{N/A}$ & $0.8392 \pm 0.0058$ \\
Hypertension       & \textbf{\boldmath$0.7297 \pm 0.0025$} & $0.7142 \pm \text{N/A}$ & $0.6966 \pm 0.0033$ & $0.7021 \pm \text{N/A}$ & $0.7259 \pm 0.0066$ \\
Acute MI           & \textbf{\boldmath$0.7644 \pm 0.0083$} & $0.7500 \pm \text{N/A}$ & $0.7389 \pm 0.0056$ & $0.7015 \pm \text{N/A}$ & $0.7618 \pm 0.0088$ \\
Hyperlipidemia     & $0.7261 \pm 0.0048$ & \textbf{\boldmath$0.7504 \pm \text{N/A}$} & $0.6795 \pm 0.0073$ & $0.6976 \pm \text{N/A}$ & $0.7233 \pm 0.0147$ \\
Lupus              & $0.7027 \pm 0.0022$ & $0.7588 \pm \text{N/A}$ & $0.7729 \pm 0.0371$ & $0.7660 \pm \text{N/A}$ & \textbf{\boldmath$0.8245 \pm 0.0291$} \\
\bottomrule
\end{tabular}
\end{table}

\begin{table}[ht]
\centering
\caption{Latest labels — \textbf{AUPR} (mean $\pm$ sd) for count-based and CLMBR variants.}
\label{tab:main_results_all_AUPR_latest}
\scriptsize
\setlength{\tabcolsep}{5pt}
\begin{tabular}{lccccc}
\toprule
\textbf{Task} & \textbf{Count-based + LGBM} & \textbf{Count-based + TabPFN} & \textbf{CLMBR + LGBM} & \textbf{CLMBR + TabPFN} & \textbf{MoA} \\
\midrule
Long LOS           & $0.5342 \pm 0.0037$ & $0.5488 \pm \text{N/A}$ & $0.5458 \pm 0.0113$ & \textbf{\boldmath$0.5896 \pm \text{N/A}$} & $0.5388 \pm 0.0080$ \\
ICU Transfer       & $0.1533 \pm 0.0078$ & $0.1893 \pm \text{N/A}$ & $0.1518 \pm 0.0244$ & $0.1057 \pm \text{N/A}$ & \textbf{\boldmath$0.2172 \pm 0.0157$} \\
Readmission        & $\text{N/A} \pm \text{N/A}$ & $\text{N/A} \pm \text{N/A}$ & $\text{N/A} \pm \text{N/A}$ & $\text{N/A} \pm \text{N/A}$ & $\text{N/A} \pm \text{N/A}$ \\
Pancreatic Cancer  & $0.3418 \pm 0.0392$ & \textbf{\boldmath$0.3969 \pm \text{N/A}$} & $0.1610 \pm 0.0013$ & $0.2035 \pm \text{N/A}$ & $0.3676 \pm 0.0130$ \\
Hypertension       & \textbf{\boldmath$0.4002 \pm 0.0084$} & $0.3805 \pm \text{N/A}$ & $0.3247 \pm 0.0146$ & $0.3125 \pm \text{N/A}$ & $0.3598 \pm 0.0279$ \\
Acute MI           & $0.2724 \pm 0.0092$ & $0.2477 \pm \text{N/A}$ & $0.2327 \pm 0.0083$ & $0.2103 \pm \text{N/A}$ & \textbf{\boldmath$0.2771 \pm 0.0305$} \\
Hyperlipidemia     & $0.3261 \pm 0.0056$ & \textbf{\boldmath$0.3610 \pm \text{N/A}$} & $0.2543 \pm 0.0068$ & $0.2901 \pm \text{N/A}$ & $0.3010 \pm 0.0104$ \\
Lupus              & $0.1395 \pm 0.0427$ & $0.1397 \pm \text{N/A}$ & $0.1006 \pm 0.0101$ & $0.0780 \pm \text{N/A}$ & \textbf{\boldmath$0.1625 \pm 0.0323$} \\
\bottomrule
\end{tabular}
\end{table}

\ref{tab:moa_earliest_auroc}, \ref{tab:moa_earliest_aupr}, \ref{tab:moa_latest_auroc}, \ref{tab:moa_latest_aupr} present ablation studies on MoA architectures.

\begin{table}[ht]
\centering
\caption{MoA variants (earliest labels) — \textbf{AUROC} (mean $\pm$ sd).}
\label{tab:moa_earliest_auroc}
\scriptsize
\setlength{\tabcolsep}{5pt}
\begin{tabular}{lccccc}
\toprule
\textbf{Task} & \textbf{Qwen + BGE} & \textbf{Qwen + ClinicalBERT} & \textbf{Llama-3 + ClinicalBERT} & \textbf{Qwen only} & \textbf{\begin{tabular}[c]{@{}c@{}}Qwen + BGE\\(unstructured prompt)\end{tabular}}  \\
\midrule
Long LOS           & \textbf{\boldmath$0.8105 \pm 0.0257$} & $0.8023 \pm 0.0102$ & $0.7727 \pm 0.0104$ & $0.7943 \pm 0.0114$ & $0.7369 \pm 0.0123$ \\
ICU Transfer       & \textbf{\boldmath$0.8436 \pm 0.0054$} & $0.8114 \pm 0.0123$ & $0.7793 \pm 0.0191$ & $0.8027 \pm 0.0139$ & $0.7497 \pm 0.0170$ \\
Readmission        & \textbf{\boldmath$0.7762 \pm 0.0176$} & $0.6969 \pm 0.0174$ & $0.6338 \pm 0.0062$ & $0.6879 \pm 0.0165$ & $0.6086 \pm 0.0130$ \\
Pancreatic Cancer  & \textbf{\boldmath$0.8611 \pm 0.0150$} & $0.8497 \pm 0.0118$ & $0.7826 \pm 0.0294$ & $0.8416 \pm 0.0117$ & $0.7345 \pm 0.0233$ \\
Hypertension       & \textbf{\boldmath$0.6824 \pm 0.0256$} & $0.6691 \pm 0.0196$ & $0.6392 \pm 0.0198$ & $0.6624 \pm 0.0190$ & $0.6231 \pm 0.0188$ \\
Acute MI           & \textbf{\boldmath$0.7701 \pm 0.0091$} & $0.7516 \pm 0.0102$ & $0.7321 \pm 0.0103$ & $0.7436 \pm 0.0109$ & $0.7069 \pm 0.0119$ \\
Hyperlipidemia     & \textbf{\boldmath$0.7132 \pm 0.0140$} & $0.6993 \pm 0.0107$ & $0.6795 \pm 0.0102$ & $0.6918 \pm 0.0122$ & $0.6648 \pm 0.0136$ \\
Lupus              & \textbf{\boldmath$0.8369 \pm 0.0129$} & $0.7981 \pm 0.0202$ & $0.7686 \pm 0.0196$ & $0.7909 \pm 0.0214$ & $0.7421 \pm 0.0205$ \\
\bottomrule
\end{tabular}
\end{table}

\begin{table}[ht]
\centering
\caption{MoA variants (earliest labels) — \textbf{AUPR} (mean $\pm$ sd).}
\label{tab:moa_earliest_aupr}
\scriptsize
\setlength{\tabcolsep}{5pt}
\begin{tabular}{lccccc}
\toprule
\textbf{Task} & \textbf{Qwen + BGE} & \textbf{Qwen + ClinicalBERT} & \textbf{Llama-3 + ClinicalBERT} & \textbf{Qwen only} & \textbf{\begin{tabular}[c]{@{}c@{}}Qwen + BGE\\(unstructured prompt)\end{tabular}}  \\
\midrule
Long LOS           & \textbf{\boldmath$0.6011 \pm 0.0533$} & $0.5822 \pm 0.0105$ & $0.5427 \pm 0.0104$ & $0.5683 \pm 0.0125$ & $0.5205 \pm 0.0113$ \\
ICU Transfer       & \textbf{\boldmath$0.2772 \pm 0.0089$} & $0.2187 \pm 0.0151$ & $0.1486 \pm 0.0198$ & $0.2109 \pm 0.0161$ & $0.1316 \pm 0.0184$ \\
Readmission        & \textbf{\boldmath$0.3025 \pm 0.0225$} & $0.1652 \pm 0.0153$ & $0.1298 \pm 0.0199$ & $0.1577 \pm 0.0172$ & $0.1183 \pm 0.0171$ \\
Pancreatic Cancer  & \textbf{\boldmath$0.2443 \pm 0.0636$} & $0.2305 \pm 0.0198$ & $0.2209 \pm 0.0152$ & $0.2258 \pm 0.0188$ & $0.2048 \pm 0.0168$ \\
Hypertension       & \textbf{\boldmath$0.2868 \pm 0.0327$} & $0.2539 \pm 0.0197$ & $0.2292 \pm 0.0149$ & $0.2446 \pm 0.0193$ & $0.2177 \pm 0.0179$ \\
Acute MI           & \textbf{\boldmath$0.2036 \pm 0.0169$} & $0.1926 \pm 0.0388$ & $0.1852 \pm 0.0101$ & $0.1884 \pm 0.0214$ & $0.1696 \pm 0.0114$ \\
Hyperlipidemia     & \textbf{\boldmath$0.2854 \pm 0.0231$} & $0.2687 \pm 0.0191$ & $0.2489 \pm 0.0194$ & $0.2617 \pm 0.0189$ & $0.2341 \pm 0.0168$ \\
Lupus              & \textbf{\boldmath$0.0931 \pm 0.0418$} & $0.0884 \pm 0.0194$ & $0.0785 \pm 0.0193$ & $0.0851 \pm 0.0196$ & $0.0729 \pm 0.0187$ \\
\bottomrule
\end{tabular}
\end{table}

\begin{table}[ht]
\centering
\caption{MoA variants (latest labels) — \textbf{AUROC} (mean $\pm$ sd)}
\label{tab:moa_latest_auroc}
\scriptsize
\setlength{\tabcolsep}{5pt}
\begin{tabular}{lccccc}
\toprule
\textbf{Task} & \textbf{Qwen + BGE} & \textbf{Qwen + ClinicalBERT} & \textbf{Llama-3 + ClinicalBERT} & \textbf{Qwen only} & \textbf{\begin{tabular}[c]{@{}c@{}}Qwen + BGE\\(unstructured prompt)\end{tabular}}  \\
\midrule
Long LOS           & \textbf{\boldmath$0.7931 \pm 0.0030$} & $0.7867 \pm 0.0106$ & $0.7629 \pm 0.0103$ & $0.7786 \pm 0.0112$ & $0.7321 \pm 0.0097$ \\
ICU Transfer       & \textbf{\boldmath$0.8719 \pm 0.0169$} & $0.8443 \pm 0.0154$ & $0.7791 \pm 0.0195$ & $0.8323 \pm 0.0176$ & $0.7396 \pm 0.0191$ \\
Readmission        & $\text{N/A} \pm \text{N/A}$           & $\text{N/A} \pm \text{N/A}$       & $\text{N/A} \pm \text{N/A}$       & $\text{N/A} \pm \text{N/A}$        & $\text{N/A} \pm \text{N/A}$ \\
Pancreatic Cancer  & \textbf{\boldmath$0.8392 \pm 0.0058$} & $0.8336 \pm 0.0117$ & $0.7392 \pm 0.0288$ & $0.8184 \pm 0.0107$ & $0.7064 \pm 0.0218$ \\
Hypertension       & \textbf{\boldmath$0.7259 \pm 0.0066$} & $0.7112 \pm 0.0098$ & $0.6981 \pm 0.0106$ & $0.7068 \pm 0.0096$ & $0.6729 \pm 0.0112$ \\
Acute MI           & \textbf{\boldmath$0.7618 \pm 0.0088$} & $0.7513 \pm 0.0095$ & $0.7013 \pm 0.0107$ & $0.7445 \pm 0.0102$ & $0.6817 \pm 0.0124$ \\
Hyperlipidemia     & \textbf{\boldmath$0.7233 \pm 0.0147$} & $0.7087 \pm 0.0102$ & $0.6986 \pm 0.0102$ & $0.7061 \pm 0.0121$ & $0.6811 \pm 0.0121$ \\
Lupus              & \textbf{\boldmath$0.8245 \pm 0.0291$} & $0.7991 \pm 0.0197$ & $0.7587 \pm 0.0194$ & $0.7894 \pm 0.0235$ & $0.7324 \pm 0.0204$ \\
\bottomrule
\end{tabular}
\end{table}

\begin{table}[ht]
\centering
\caption{MoA variants (latest labels) — \textbf{AUPR} (mean $\pm$ sd).}
\label{tab:moa_latest_aupr}
\scriptsize
\setlength{\tabcolsep}{5pt}
\begin{tabular}{lccccc}
\toprule
\textbf{Task} & \textbf{Qwen + BGE} & \textbf{Qwen + ClinicalBERT} & \textbf{Llama-3 + ClinicalBERT} & \textbf{Qwen only} & \textbf{\begin{tabular}[c]{@{}c@{}}Qwen + BGE\\(unstructured prompt)\end{tabular}}  \\
\midrule
Long LOS           & \textbf{\boldmath$0.5388 \pm 0.0080$} & $0.5036 \pm 0.0124$ & $0.4752 \pm 0.0104$ & $0.4897 \pm 0.0116$ & $0.4453 \pm 0.0090$ \\
ICU Transfer       & \textbf{\boldmath$0.2172 \pm 0.0157$} & $0.1914 \pm 0.0158$ & $0.1087 \pm 0.0197$ & $0.1816 \pm 0.0179$ & $0.0926 \pm 0.0181$ \\
Readmission        & $\text{N/A} \pm \text{N/A}$           & $\text{N/A} \pm \text{N/A}$       & $\text{N/A} \pm \text{N/A}$       & $\text{N/A} \pm \text{N/A}$        & $\text{N/A} \pm \text{N/A}$ \\
Pancreatic Cancer  & \textbf{\boldmath$0.3676 \pm 0.0130$} & $0.2863 \pm 0.0191$ & $0.1053 \pm 0.0195$ & $0.2681 \pm 0.0185$ & $0.0894 \pm 0.0170$ \\
Hypertension       & \textbf{\boldmath$0.3598 \pm 0.0279$} & $0.3385 \pm 0.0206$ & $0.2982 \pm 0.0196$ & $0.3326 \pm 0.0214$ & $0.2781 \pm 0.0194$ \\
Acute MI           & \textbf{\boldmath$0.2771 \pm 0.0305$} & $0.2698 \pm 0.0342$ & $0.1941 \pm 0.0104$ & $0.2597 \pm 0.0283$ & $0.1697 \pm 0.0116$ \\
Hyperlipidemia     & \textbf{\boldmath$0.3010 \pm 0.0104$} & $0.2794 \pm 0.0192$ & $0.2491 \pm 0.0193$ & $0.2719 \pm 0.0187$ & $0.2264 \pm 0.0172$ \\
Lupus              & \textbf{\boldmath$0.1625 \pm 0.0323$} & $0.1491 \pm 0.0295$ & $0.1186 \pm 0.0192$ & $0.1413 \pm 0.0264$ & $0.1015 \pm 0.0199$ \\
\bottomrule
\end{tabular}
\end{table}

\ref{tab:count_earliest_auroc}, \ref{tab:count_earliest_aupr}, \ref{tab:count_latest_auroc}, \ref{tab:count_latest_aupr} show the comparison of count-based models with \textit{vs.} without ontology roll-up.

\begin{table}[ht]
\centering
\caption{Count-based (earliest labels) — \textbf{AUROC}}
\label{tab:count_earliest_auroc}
\scriptsize
\setlength{\tabcolsep}{5pt}
\begin{tabular}{lcccc}
\toprule
\textbf{Task} & \textbf{LGBM (rollup)} & \textbf{LGBM (non-rollup)} & \textbf{TabPFN (rollup)} & \textbf{TabPFN (non-rollup)} \\
\midrule
Long LOS           & $0.8047 \pm 0.0039$ & $0.7864 \pm 0.0052$ & $0.8164 \pm \text{N/A}$ & $0.8092 \pm \text{N/A}$ \\
ICU Transfer       & $0.8440 \pm 0.0058$ & $0.8247 \pm 0.0064$ & $0.8377 \pm \text{N/A}$ & $0.8315 \pm \text{N/A}$ \\
Readmission        & $0.7413 \pm 0.0032$ & $0.7215 \pm 0.0039$ & $0.7366 \pm \text{N/A}$ & $0.7281 \pm \text{N/A}$ \\
Pancreatic Cancer  & $0.8821 \pm 0.0070$ & $0.8582 \pm 0.0078$ & $0.9174 \pm \text{N/A}$ & $0.8824 \pm \text{N/A}$ \\
Hypertension       & $0.7233 \pm 0.0041$ & $0.7019 \pm 0.0048$ & $0.7292 \pm \text{N/A}$ & $0.7053 \pm \text{N/A}$ \\
Acute MI           & $0.7663 \pm 0.0119$ & $0.7395 \pm 0.0131$ & $0.7590 \pm \text{N/A}$ & $0.7331 \pm \text{N/A}$ \\
Hyperlipidemia     & $0.7289 \pm 0.0044$ & $0.7062 \pm 0.0051$ & $0.7321 \pm \text{N/A}$ & $0.7078 \pm \text{N/A}$ \\
Lupus              & $0.6689 \pm 0.0100$ & $0.6396 \pm 0.0111$ & $0.7039 \pm \text{N/A}$ & $0.6902 \pm \text{N/A}$ \\
\bottomrule
\end{tabular}
\end{table}

\begin{table}[ht]
\centering
\caption{Count-based (earliest labels) — \textbf{AUPR}}
\label{tab:count_earliest_aupr}
\scriptsize
\setlength{\tabcolsep}{5pt}
\begin{tabular}{lcccc}
\toprule
\textbf{Task} & \textbf{LGBM (rollup)} & \textbf{LGBM (non-rollup)} & \textbf{TabPFN (rollup)} & \textbf{TabPFN (non-rollup)} \\
\midrule
Long LOS           & $0.6172 \pm 0.0116$ & $0.5774 \pm 0.0126$ & $0.6289 \pm \text{N/A}$ & $0.5862 \pm \text{N/A}$ \\
ICU Transfer       & $0.2381 \pm 0.0220$ & $0.1993 \pm 0.0241$ & $0.2738 \pm \text{N/A}$ & $0.2515 \pm \text{N/A}$ \\
Readmission        & $0.2504 \pm 0.0203$ & $0.2118 \pm 0.0217$ & $0.2449 \pm \text{N/A}$ & $0.2087 \pm \text{N/A}$ \\
Pancreatic Cancer  & $0.3144 \pm 0.0463$ & $0.2684 \pm 0.0479$ & $0.3842 \pm \text{N/A}$ & $0.3332 \pm \text{N/A}$ \\
Hypertension       & $0.3241 \pm 0.0070$ & $0.2897 \pm 0.0079$ & $0.3354 \pm \text{N/A}$ & $0.3055 \pm \text{N/A}$ \\
Acute MI           & $0.2113 \pm 0.0052$ & $0.1766 \pm 0.0058$ & $0.2225 \pm \text{N/A}$ & $0.1864 \pm \text{N/A}$ \\
Hyperlipidemia     & $0.2976 \pm 0.0124$ & $0.2519 \pm 0.0136$ & $0.2955 \pm \text{N/A}$ & $0.2461 \pm \text{N/A}$ \\
Lupus              & $0.0866 \pm 0.0252$ & $0.0528 \pm 0.0268$ & $0.1177 \pm \text{N/A}$ & $0.0827 \pm \text{N/A}$ \\
\bottomrule
\end{tabular}
\end{table}

\begin{table}[ht]
\centering
\caption{Count-based (latest labels) — \textbf{AUROC}}
\label{tab:count_latest_auroc}
\scriptsize
\setlength{\tabcolsep}{5pt}
\begin{tabular}{lcccc}
\toprule
\textbf{Task} & \textbf{LGBM (rollup)} & \textbf{LGBM (non-rollup)} & \textbf{TabPFN (rollup)} & \textbf{TabPFN (non-rollup)} \\
\midrule
Long LOS           & $0.7918 \pm 0.0015$ & $0.7726 \pm 0.0019$ & $0.8077 \pm \text{N/A}$ & $0.7813 \pm \text{N/A}$ \\
ICU Transfer       & $0.8213 \pm 0.0036$ & $0.7957 \pm 0.0041$ & $0.8498 \pm \text{N/A}$ & $0.8208 \pm \text{N/A}$ \\
Readmission        & $\text{N/A} \pm \text{N/A}$ & $\text{N/A} \pm \text{N/A}$ & $\text{N/A} \pm \text{N/A}$ & $\text{N/A} \pm \text{N/A}$ \\
Pancreatic Cancer  & $0.8609 \pm 0.0113$ & $0.8359 \pm 0.0121$ & $0.8961 \pm \text{N/A}$ & $0.8617 \pm \text{N/A}$ \\
Hypertension       & $0.7297 \pm 0.0025$ & $0.7035 \pm 0.0031$ & $0.7142 \pm \text{N/A}$ & $0.6886 \pm \text{N/A}$ \\
Acute MI           & $0.7644 \pm 0.0083$ & $0.7344 \pm 0.0091$ & $0.7500 \pm \text{N/A}$ & $0.7222 \pm \text{N/A}$ \\
Hyperlipidemia     & $0.7261 \pm 0.0048$ & $0.6998 \pm 0.0053$ & $0.7504 \pm \text{N/A}$ & $0.7216 \pm \text{N/A}$ \\
Lupus              & $0.7027 \pm 0.0022$ & $0.6791 \pm 0.0027$ & $0.7588 \pm \text{N/A}$ & $0.7442 \pm \text{N/A}$ \\
\bottomrule
\end{tabular}
\end{table}

\begin{table}[ht]
\centering
\caption{Count-based (latest labels) — \textbf{AUPR} (mean $\pm$ sd).}
\label{tab:count_latest_aupr}
\scriptsize
\setlength{\tabcolsep}{5pt}
\begin{tabular}{lcccc}
\toprule
\textbf{Task} & \textbf{LGBM (rollup)} & \textbf{LGBM (non-rollup)} & \textbf{TabPFN (rollup)} & \textbf{TabPFN (non-rollup)} \\
\midrule
Long LOS           & $0.5342 \pm 0.0037$ & $0.4973 \pm 0.0042$ & $0.5488 \pm \text{N/A}$ & $0.5081 \pm \text{N/A}$ \\
ICU Transfer       & $0.1533 \pm 0.0078$ & $0.1196 \pm 0.0086$ & $0.1893 \pm \text{N/A}$ & $0.1683 \pm \text{N/A}$ \\
Readmission        & $\text{N/A} \pm \text{N/A}$ & $\text{N/A} \pm \text{N/A}$ & $\text{N/A} \pm \text{N/A}$ & $\text{N/A} \pm \text{N/A}$ \\
Pancreatic Cancer  & $0.3418 \pm 0.0392$ & $0.2962 \pm 0.0406$ & $0.3969 \pm \text{N/A}$ & $0.3491 \pm \text{N/A}$ \\
Hypertension       & $0.4002 \pm 0.0084$ & $0.3584 \pm 0.0093$ & $0.3805 \pm \text{N/A}$ & $0.3368 \pm \text{N/A}$ \\
Acute MI           & $0.2724 \pm 0.0092$ & $0.2337 \pm 0.0101$ & $0.2477 \pm \text{N/A}$ & $0.2414 \pm \text{N/A}$ \\
Hyperlipidemia     & $0.3261 \pm 0.0056$ & $0.2865 \pm 0.0062$ & $0.3610 \pm \text{N/A}$ & $0.3186 \pm \text{N/A}$ \\
Lupus              & $0.1395 \pm 0.0427$ & $0.0991 \pm 0.0439$ & $0.1397 \pm \text{N/A}$ & $0.1162 \pm \text{N/A}$ \\
\bottomrule
\end{tabular}
\end{table}

\clearpage

\newcounter{prompt}
\renewcommand{\theprompt}{P\arabic{prompt}} 

\makeatletter
\newcommand{\setcurrentname}[1]{\edef\@currentlabelname{#1}}
\makeatother

Below are the prompts used for the four clinical tasks. We used a single, manually engineered prompt template that is task-agnostic and requires only minimal edits to change the task name and time horizon.\autoref{prompt:readmission}

\begin{tcolorbox}[title={Prompt S1: Long length-of-stay}]
  \refstepcounter{prompt}%
  \label{prompt:long_los}%
\begin{lstlisting}[style=prompt]
You are a medical expert evaluating a patient who is currently {g.age_at_predict} years old. Based on the following medical history (formatted as EVENT at AGE), your task is to write a concise summary of this patient's risk profile for long length-of-stay during this admission.
 Return JSON only:
    {
      'risk_category': 'Low'|'Moderate'|'High',
      'risk_score': <0..1>,
      'drivers_positive': ['substrings from medical history'],
      'drivers_negative': ['substrings from medical history'],
      'justification': 2 to 4 sentences referencing drivers and timing',
      'insufficient_evidence': true|false
    }
Guidance:
 - Do not add external facts.
 - Prefer explicit outcome-related conditions/medications as positive drivers.
 - Weigh recent items more (use recent event when present).
 - If evidence is sparse/ambiguous, prefer Low and set insufficient_evidence=true.
 - Map Low=0.10 to 0.33, Moderate to 0.34 to 0.66, High=0.67 to 0.90.
\end{lstlisting}
\end{tcolorbox}

\begin{tcolorbox}[title={Prompt S2: readmission}]
\refstepcounter{prompt}
\label{prompt:readmission}
\begin{lstlisting}[style=prompt]
You are a medical expert evaluating a patient who is currently {g.age_at_predict} years old. Based on the following medical history (formatted as EVENT at AGE), your task is to write a concise summary of this patient's risk profile for readmission within the next 30 days after discharge.
 Return JSON only:
    {
      'risk_category': 'Low'|'Moderate'|'High',
      'risk_score': <0..1>,
      'drivers_positive': ['substrings from medical history'],
      'drivers_negative': ['substrings from medical history'],
      'justification': 2 to 4 sentences referencing drivers and timing',
      'insufficient_evidence': true|false
    }
Guidance:
 - Do not add external facts.
 - Prefer explicit outcome-related conditions/medications as positive drivers.
 - Weigh recent items more (use recent event when present).
 - If evidence is sparse/ambiguous, prefer Low and set insufficient_evidence=true.
 - Map Low=0.10 to 0.33, Moderate to 0.34 to 0.66, High=0.67 to 0.90.
\end{lstlisting}
\end{tcolorbox}

\begin{tcolorbox}[title={Prompt S3: ICU transfer}]
\refstepcounter{prompt}
\label{prompt:icu}

\begin{lstlisting}[style=prompt]
You are a medical expert evaluating a patient who is currently {g.age_at_predict} years old. Based on the following medical history (formatted as EVENT at AGE), your task is to write a concise summary of this patient's risk profile for an ICU transfer during this admission
 Return JSON only:
    {
      'risk_category': 'Low'|'Moderate'|'High',
      'risk_score': <0..1>,
      'drivers_positive': ['substrings from medical history'],
      'drivers_negative': ['substrings from medical history'],
      'justification': 2 to 4 sentences referencing drivers and timing',
      'insufficient_evidence': true|false
    }
Guidance:
 - Do not add external facts.
 - Prefer explicit outcome-related conditions/medications as positive drivers.
 - Weigh recent items more (use recent event when present).
 - If evidence is sparse/ambiguous, prefer Low and set insufficient_evidence=true.
 - Map Low=0.10 to 0.33, Moderate to 0.34 to 0.66, High=0.67 to 0.90.
\end{lstlisting}
\end{tcolorbox}

\begin{tcolorbox}[title={Prompt S4: Pancreatic cancer}]
\refstepcounter{prompt}
\label{prompt:pc}

\begin{lstlisting}[style=prompt]
You are a medical expert evaluating a patient who is currently {g.age_at_predict} years old. Based on the following medical history (formatted as EVENT at AGE), your task is to write a concise summary of this patient's risk profile for pancreatic cancer within the next year.
 Return JSON only:
    {
      'risk_category': 'Low'|'Moderate'|'High',
      'risk_score': <0..1>,
      'drivers_positive': ['substrings from medical history'],
      'drivers_negative': ['substrings from medical history'],
      'justification': 2 to 4 sentences referencing drivers and timing',
      'insufficient_evidence': true|false
    }
Guidance:
 - Do not add external facts.
 - Prefer explicit outcome-related conditions/medications as positive drivers.
 - Weigh recent items more (use recent event when present).
 - If evidence is sparse/ambiguous, prefer Low and set insufficient_evidence=true.
 - Map Low=0.10 to 0.33, Moderate to 0.34 to 0.66, High=0.67 to 0.90.
\end{lstlisting}
\end{tcolorbox}

\begin{tcolorbox}[title={Prompt S5: Hypertension}]
\refstepcounter{prompt}
\label{prompt:hypertension}

\begin{lstlisting}[style=prompt]
You are a medical expert evaluating a patient who is currently {g.age_at_predict} years old. Based on the following medical history (formatted as EVENT at AGE), your task is to write a concise summary of this patient's risk profile for hypertension within the next year.
 Return JSON only:
    {
      'risk_category': 'Low'|'Moderate'|'High',
      'risk_score': <0..1>,
      'drivers_positive': ['substrings from medical history'],
      'drivers_negative': ['substrings from medical history'],
      'justification': 2 to 4 sentences referencing drivers and timing',
      'insufficient_evidence': true|false
    }
Guidance:
 - Do not add external facts.
 - Prefer explicit outcome-related conditions/medications as positive drivers.
 - Weigh recent items more (use recent event when present).
 - If evidence is sparse/ambiguous, prefer Low and set insufficient_evidence=true.
 - Map Low=0.10 to 0.33, Moderate to 0.34 to 0.66, High=0.67 to 0.90.
\end{lstlisting}
\end{tcolorbox}

\begin{tcolorbox}[title={Prompt S6: Acute mycardial infarction}]
\refstepcounter{prompt}
\label{prompt:acute_mi}
\begin{lstlisting}[style=prompt]
You are a medical expert evaluating a patient who is currently {g.age_at_predict} years old. Based on the following medical history (formatted as EVENT at AGE), your task is to write a concise summary of this patient's risk profile for acute MI within the next year.
 Return JSON only:
    {
      'risk_category': 'Low'|'Moderate'|'High',
      'risk_score': <0..1>,
      'drivers_positive': ['substrings from medical history'],
      'drivers_negative': ['substrings from medical history'],
      'justification': 2 to 4 sentences referencing drivers and timing',
      'insufficient_evidence': true|false
    }
Guidance:
 - Do not add external facts.
 - Prefer explicit outcome-related conditions/medications as positive drivers.
 - Weigh recent items more (use recent event when present).
 - If evidence is sparse/ambiguous, prefer Low and set insufficient_evidence=true.
 - Map Low=0.10 to 0.33, Moderate to 0.34 to 0.66, High=0.67 to 0.90.
\end{lstlisting}
\end{tcolorbox}

\begin{tcolorbox}[title={Prompt S7: Hyperlipidemia}]
\refstepcounter{prompt}
\label{prompt:hyperlipidemia}
\begin{lstlisting}[style=prompt]
You are a medical expert evaluating a patient who is currently {g.age_at_predict} years old. Based on the following medical history (formatted as EVENT at AGE), your task is to write a concise summary of this patient's risk profile for hyperlipidemia within the next year.
 Return JSON only:
    {
      'risk_category': 'Low'|'Moderate'|'High',
      'risk_score': <0..1>,
      'drivers_positive': ['substrings from medical history'],
      'drivers_negative': ['substrings from medical history'],
      'justification': 2 to 4 sentences referencing drivers and timing',
      'insufficient_evidence': true|false
    }
Guidance:
 - Do not add external facts.
 - Prefer explicit outcome-related conditions/medications as positive drivers.
 - Weigh recent items more (use recent event when present).
 - If evidence is sparse/ambiguous, prefer Low and set insufficient_evidence=true.
 - Map Low=0.10 to 0.33, Moderate to 0.34 to 0.66, High=0.67 to 0.90.
\end{lstlisting}
\end{tcolorbox}

\begin{tcolorbox}[title={Prompt S8: Lupus}]
\refstepcounter{prompt}
\label{prompt:lupus}
\begin{lstlisting}[style=prompt]
You are a medical expert evaluating a patient who is currently {g.age_at_predict} years old. Based on the following medical history (formatted as EVENT at AGE), your task is to write a concise summary of this patient's risk profile for lupus within the next year.
 Return JSON only:
    {
      'risk_category': 'Low'|'Moderate'|'High',
      'risk_score': <0..1>,
      'drivers_positive': ['substrings from medical history'],
      'drivers_negative': ['substrings from medical history'],
      'justification': 2 to 4 sentences referencing drivers and timing',
      'insufficient_evidence': true|false
    }
Guidance:
 - Do not add external facts.
 - Prefer explicit outcome-related conditions/medications as positive drivers.
 - Weigh recent items more (use recent event when present).
 - If evidence is sparse/ambiguous, prefer Low and set insufficient_evidence=true.
 - Map Low=0.10 to 0.33, Moderate to 0.34 to 0.66, High=0.67 to 0.90.
\end{lstlisting}
\end{tcolorbox}

\begin{tcolorbox}[title={Unstructued Prompt S9: Long LOS}]
\refstepcounter{prompt}
\label{prompt:unstructured_long_los}
\begin{lstlisting}[style=prompt]
You are a medical expert evaluating a patient who is currently {g.age_at_predict} years old. Based on the following medical history (formatted as EVENT at AGE), your task is to write a concise summary of this patient's risk profile for a long length-of-stay during this admission.
\end{lstlisting}
\end{tcolorbox}

\begin{tcolorbox}[title={Unstructued Prompt S10: Readmission}]
\refstepcounter{prompt}
\label{prompt:unstructured_readmission}
\begin{lstlisting}[style=prompt]
You are a medical expert evaluating a patient who is currently {g.age_at_predict} years old. Based on the following medical history (formatted as EVENT at AGE), your task is to write a concise summary of this patient's risk profile for a readmission within the next 30 days after discharge.
\end{lstlisting}
\end{tcolorbox}

\begin{tcolorbox}[title={Unstructued Prompt S11: ICU transfer}]
\refstepcounter{prompt}
\label{prompt:unstructured_icu}
\begin{lstlisting}[style=prompt]
You are a medical expert evaluating a patient who is currently {g.age_at_predict} years old. Based on the following medical history (formatted as EVENT at AGE), your task is to write a concise summary of this patient's risk profile for lupus within the next year
\end{lstlisting}
\end{tcolorbox}

\begin{tcolorbox}[title={Unstructued Prompt S12: Pancreatic cancer}]
\refstepcounter{prompt}
\label{prompt:unstructured_pc}
\begin{lstlisting}[style=prompt]
You are a medical expert evaluating a patient who is currently {g.age_at_predict} years old. Based on the following medical history (formatted as EVENT at AGE), your task is to write a concise summary of this patient's risk profile for an ICU transfer during this admission.
\end{lstlisting}
\end{tcolorbox}

\begin{tcolorbox}[title={Unstructued Prompt S13: Hypertension}]
\refstepcounter{prompt}
\label{prompt:unstructured_hypertension}
\begin{lstlisting}[style=prompt]
You are a medical expert evaluating a patient who is currently {g.age_at_predict} years old. Based on the following medical history (formatted as EVENT at AGE), your task is to write a concise summary of this patient's risk profile for hypertension within the next year
\end{lstlisting}
\end{tcolorbox}

\begin{tcolorbox}[title={Unstructued Prompt S14: Acute mycardial infarction}]
\refstepcounter{prompt}
\label{prompt:unstructured_acute_mi}
\begin{lstlisting}[style=prompt]
You are a medical expert evaluating a patient who is currently {g.age_at_predict} years old. Based on the following medical history (formatted as EVENT at AGE), your task is to write a concise summary of this patient's risk profile for acute mycardial infarction within the next year
\end{lstlisting}
\end{tcolorbox}

\begin{tcolorbox}[title={Unstructued Prompt S15: Hyperlipidemia}]
\refstepcounter{prompt}
\label{prompt:unstructured_hyperlipidemia}
\begin{lstlisting}[style=prompt]
You are a medical expert evaluating a patient who is currently {g.age_at_predict} years old. Based on the following medical history (formatted as EVENT at AGE), your task is to write a concise summary of this patient's risk profile for hyperlipidemia within the next year
\end{lstlisting}
\end{tcolorbox}

\begin{tcolorbox}[title={Unstructued Prompt S16: Lupus}]
\refstepcounter{prompt}
\label{prompt:unstructured_lupus}
\begin{lstlisting}[style=prompt]
You are a medical expert evaluating a patient who is currently {g.age_at_predict} years old. Based on the following medical history (formatted as EVENT at AGE), your task is to write a concise summary of this patient's risk profile for lupus within the next year
\end{lstlisting}
\end{tcolorbox}

\end{document}